\documentclass{llncs}

\usepackage{llncsdoc}

\usepackage[utf8x]{inputenc} 
\usepackage[T1]{fontenc}    
\usepackage{makecell}
\usepackage{appendix}
\usepackage{enumitem}
\usepackage{graphicx}
\usepackage{caption}
\usepackage{subcaption}
\usepackage{hyperref}       
\usepackage{url}            
\usepackage{booktabs}       
\usepackage{amsfonts}       
\usepackage{nicefrac}       
\usepackage{microtype}      
\usepackage{amsmath}
\usepackage{multirow}

\title{MRNet-Product2Vec: A Multi-task Recurrent Neural Network for Product Embeddings}

\author{%
Arijit Biswas, 
Mukul Bhutani and
Subhajit Sanyal
}%
\institute{
Core Machine Learning, Amazon, Bangalore, India\\
\email{barijit,mbhutani,subhajs@amazon.com}\\ 
}

\begin{document}

\maketitle

\begin{abstract}

E-commerce websites such as Amazon, Alibaba, Flipkart, and Walmart sell billions of products. Machine learning (ML) algorithms involving products are often used to improve the customer experience and increase revenue, e.g., product similarity, recommendation, and price estimation. The products are required to be represented as features before training an ML algorithm. In this paper, we propose an approach called MRNet-Product2Vec for creating generic embeddings of products within an e-commerce ecosystem. We learn a dense and low-dimensional embedding where a diverse set of signals related to a product are explicitly injected into its representation. We train a Discriminative Multi-task Bidirectional Recurrent Neural Network (RNN), where the input is a product title fed through a Bidirectional RNN and at the output, product labels corresponding to fifteen different tasks are predicted. The task set includes several intrinsic characteristics about a product such as price, weight, size, color, popularity, and material. We evaluate the proposed embedding quantitatively and qualitatively. We demonstrate that they are almost as good as sparse and extremely high-dimensional TF-IDF representation in spite of having less than 3\% of the TF-IDF dimension. We also use a multimodal autoencoder for comparing products from different language-regions and show preliminary yet promising qualitative results.

\end{abstract}

\section{Introduction}

Large e-commerce companies such as Amazon, Alibaba, Flipkart, and Walmart sell billions of products through their websites. Data scientists across these companies try to solve hundreds of machine learning (ML) problems everyday that involve products, e.g., duplicate product detection, product recommendation, safety classification, and price estimation. The first step towards training any ML model usually involves creating feature representation of the relevant entities, i.e., products in this scenario. However, searching through hundreds of data resources within a company, identifying the relevant information, processing and transforming a product related data to a feature vector is a tedious and time consuming process. Furthermore, teams of data scientists performing such tasks on a regular basis makes the overall process inefficient and wasteful.

For typical ML tasks such as classification, regression, and similarity retrieval, a product can be represented in several ways. One of the most common approaches to represent a product is using an order-independent bag-of-words approach leveraging the textual metadata associated with the product. In this approach, one constructs a TF-IDF vector representation based on the title, description, and bullet points of a product. Although these representations are effective as features in a wide variety of classification tasks, they are usually high-dimensional and sparse, e.g., a TF-IDF representation with only 300K product titles and a minimum document frequency of 5 represents each product using more than 20K dimensions, where typically only 0.05\% of the features are non-zero. Using these high-dimensional features creates several problems in practice\footnote{Curse of dimensionality~\cite{keogh2011curse}}: (a) overfitting, i.e., does not generalize to novel test data, (b) training ML algorithms using these high-dimensional features is usually computational and storage inefficient, (c) computing semantically meaningful nearest neighbors is not straightforward and (d) they cannot be directly used in downstream ML algorithms such as Deep Neural Networks (DNN) as that increases the number of parameters significantly. On the other hand, using dense and low-dimensional features could alleviate these issues. In this paper, our goal is to create a generic, low-dimensional and dense product representation which can work almost as effectively as the high-dimensional TF-IDF representation.

We propose a novel discriminative Multi-task Learning Framework where we inject different kinds of signals pertaining to a product into its embedding. These signals capture different static aspects, such as color, material, weight, size, subcategory, target-gender, and dynamic aspects such as price, popularity, and views of a product. Each signal is captured by formulating a classification/regression or decoding task depending on the type of the corresponding label. The proposed architecture contains a Bidirectional Recurrent Neural Network (RNN) with LSTM cells as the input layer which takes the sequence of words in a product title as input and creates a hidden representation, which we refer to as ``product embeddings''. During training phase, the embeddings are fed into multiple classification/regression/decoding units corresponding to the training tasks. The full multi-task network is trained jointly in an end-to-end manner. We refer to the proposed approach as MRNet (Discriminative \underline{M}ulti-task \underline{R}ecurrent Neural \underline{Net}work) and the embeddings created using this method are referred to as MRNet-Product2Vec. Section \ref{proposed_approach} elaborates more on this.

Products sold on e-commerce websites usually belong to multiple Product Groups (PG) such as furniture, jewelry, clothes, books, home, and sports items. Some of the signals which we inject within products are PG-specific. For example, the weights of home items have a very different distribution than the weights of jewelry. Similarly, sizes of clothes (L, XL, XXL etc.) could be quite different from the sizes of furniture (king, queen etc.). We believe that a common embedding for all products across all PGs will not be able to capture the intra-PG variations. Hence, we initially learn the embeddings in a PG-specific manner and then use a sparse autoencoder to project the PG-specific embeddings to a PG-agnostic space. This ensures that MRNet-Product2Vec can also be used when the train or test data for an ML model belong to multiple PGs. Section \ref{gl_agnostic} provides more details on this. 

We encode different signals about products in the embeddings such that the embeddings are as generic as possible. However, creating embeddings that will work well for every product related ML task without further feature processing is not easy and perhaps impossible. So, we create these embeddings keeping two particular e-commerce use-cases in mind: (a) Anyone building an ML model with products can use these embeddings to build a good baseline model with little effort. (b) Someone who has a set of task specific features can use these embeddings as a means to augment with the generic signals captured in these representations. Our end-goal is to provide a generic feature representation for each product in an e-commerce system, such that data scientists don't have to spend days or months to build their first prototype. 

We evaluate MRNet-Product2Vec in both quantitative and qualitative ways. MRNet-Product2Vec is applied to five different classification tasks: (i) plugs, (ii) Ship In Its Own Container (SIOC), (iii) browse category, (iv) ingestible and (v) SIOC with unseen population (Sect. \ref{quantitative_analysis}). We compare MRNet-Product2Vec with a TF-IDF bag-of-words (sparse high-dimensional) on title words and show that in spite of having a much lower dimension than TF-IDF, MRNet-Product2Vec is comparable to TF-IDF representation. It performs almost as good as TF-IDF in two of these tasks, better than TF-IDF in two of these tasks and worse than TF-IDF in the remaining task. In Sect. \ref{qualitative_analysis} and \ref{feat_interpret}, we provide the qualitative analysis of MRNet-Prod2Vec. In Sect. \ref{language_agnostic}, we use a variant of multimodal autoencoder~\cite{ngiam2011multimodal} that can be used to compare products sold in different language-regions/countries. Preliminary qualitative results using this approach are also provided. 

\section{Prior Work}

There have been several prior works on entity embeddings using deep neural networks. Perhaps the most famous work on entity embeddings is the word2vec method~\cite{word2vec}, where continuous and distributed vector representations of words are learned based on their co-occurrences in a large text corpus. There are also a few prior research works for creating product embeddings for recommendation. Prod2Vec~\cite{prod2vec_kdd_2015} uses a word2vec-like approach that learns vector representations of products from email receipt logs by using a notion of a purchase sequence as a ``sentence'' and products within the sequence as ``words''. The product representations are used for recommendation. The authors in \cite{meta_prod2vec_recsys_2016} propose Meta-Prod2Vec, which extends the Prod2Vec~\cite{prod2vec_kdd_2015} loss by including additional interaction terms involving products' meta-data. However, these embeddings are specifically fine-tuned for a predefined end-task, i.e., recommendation and may not perform well on a wide variety of product related ML tasks.

Traditionally multi-task learning has been used when one or all of the individual tasks have smaller training datasets and the tasks are somehow correlated \cite{caruana1998multitask}. The training data from other correlated tasks should improve the learning of a particular task. However, we do not have any paucity of data and the tasks which are used to train MRNet-Product2Vec are largely uncorrelated. We have used ``unrelated'' multi-task learning such that the learned representations are generic. To the best of our knowledge, this is the first work, that performs multi-task learning in an RNN to explicitly encode different kinds of static and dynamic signals for a generic entity embedding.

\section{Proposed Approach}
\label{proposed_approach}

In this section, we describe the proposed embedding MRNet-Product2Vec. In MRNet-Product2Vec, we feed the vector representation of each word in a product title to a Bidirectional RNN. We use word2vec~\cite{word2vec} to create a dense and compact representation of all words in the product catalog. A large corpus of text is created comprising the titles and descriptions of 143 million randomly selected products from the catalog. We use Gensim\footnote{https://radimrehurek.com/gensim/} to learn a 128 dimensional word2vec representation of all the words in the corpus which occur at least 10 times. 

\subsection{MRNet-Product2Vec}
\label{MRNet-ASIN2Vec}

The proposed embeddings MRNet-Product2Vec are created by explicitly introducing different kinds of static and dynamic signals into the embeddings using a Discriminative Multi-task Bidirectional RNN. The goal of injecting different signals is to create embeddings which are as generic as possible. We believe that the learned embeddings will be effective in any ML task, which is correlated with one or more of the tasks for which we train our embeddings (see Sect. \ref{feat_interpret}).

We describe fifteen different tasks which are used to learn our product embeddings. These tasks were selected primarily because we thought that the corresponding signals are intrinsic and should be included in a generic product embedding. However, these set of tasks are not exhaustive and may not capture all possible information about the products. Future research could incorporate more tasks during training and also study if dense product embeddings of a small fixed dimension (say, 128 or 256) can capture more signals effectively.

The set of present tasks can be grouped in several ways. Some of these capture static information that are unlikely to change over the lifetime of a product, e.g., size, weight, and material. Some are also dynamic which are likely to change every week or month, e.g., price or number of views. Some of these tasks are classification problems, where some others are regression or decoding. We summarize all the tasks in Table \ref{tasks} and omit the details due to lack of space. Color, size, material, subcategory, and item type are formulated as multi-class classification problems where the most frequent ones are treated as individual classes and the remaining ones are grouped as one class. Rest of the classification tasks are binary. As mentioned earlier, this list of tasks may not be exhaustive. However, they capture a wide variety of aspects regarding a product and effective encoding of these signals should create embeddings that are generic enough to address a wide class of ML problems pertaining to products.

\begin{center}
\scriptsize
\captionof{table}{Tasks used to train MRNet-Product2Vec.}
\label{tasks}
  \begin{tabular}{ | c | c | c |}
    \hline
      & Static & Dynamic \\ \hline
    Classification  & \makecell{{\bf Color, Size, Material,}\\ {\bf Subcategory, Item Type, Hazardous,} \\ {\bf  Batteries, High Value, Target Gender}} & {\bf Offer, Review}   \\ \hline
    Regression & {\bf Weight}  & {\bf Price, View Count}  \\ \hline
    Decoding & {\bf TF-IDF representation (5000 dim.)}  & \\
    \hline
  \end{tabular}
\end{center}

\begin{figure}
    \centering
    \begin{subfigure}[b]{0.49\textwidth}
        \includegraphics[width=\textwidth]{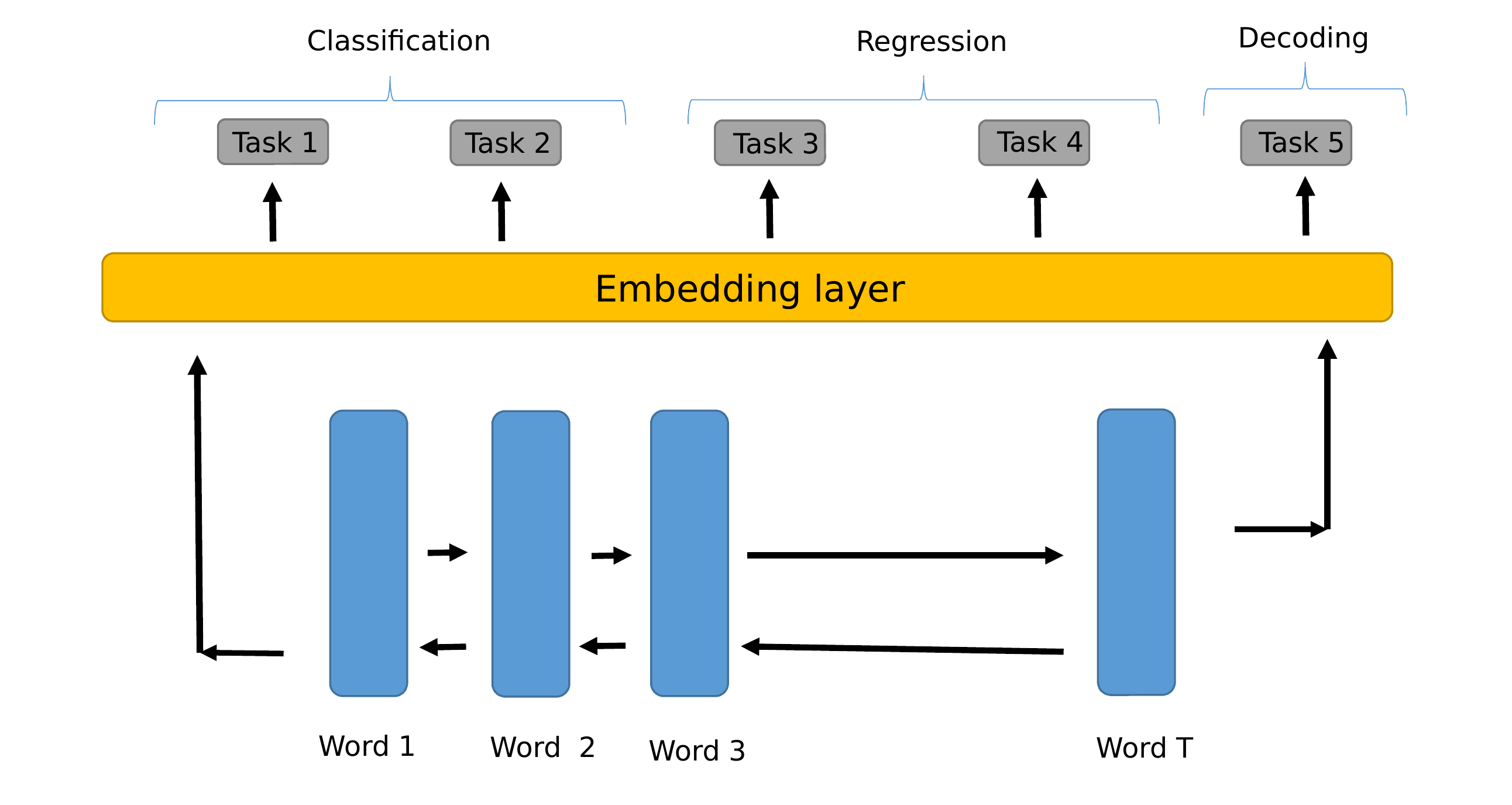}
        \caption{MRNet-Product2Vec for PG-specific embeddings.}
        \label{arch_1}
    \end{subfigure}
    ~ 
    \begin{subfigure}[b]{0.48\textwidth}
        \includegraphics[width=\textwidth]{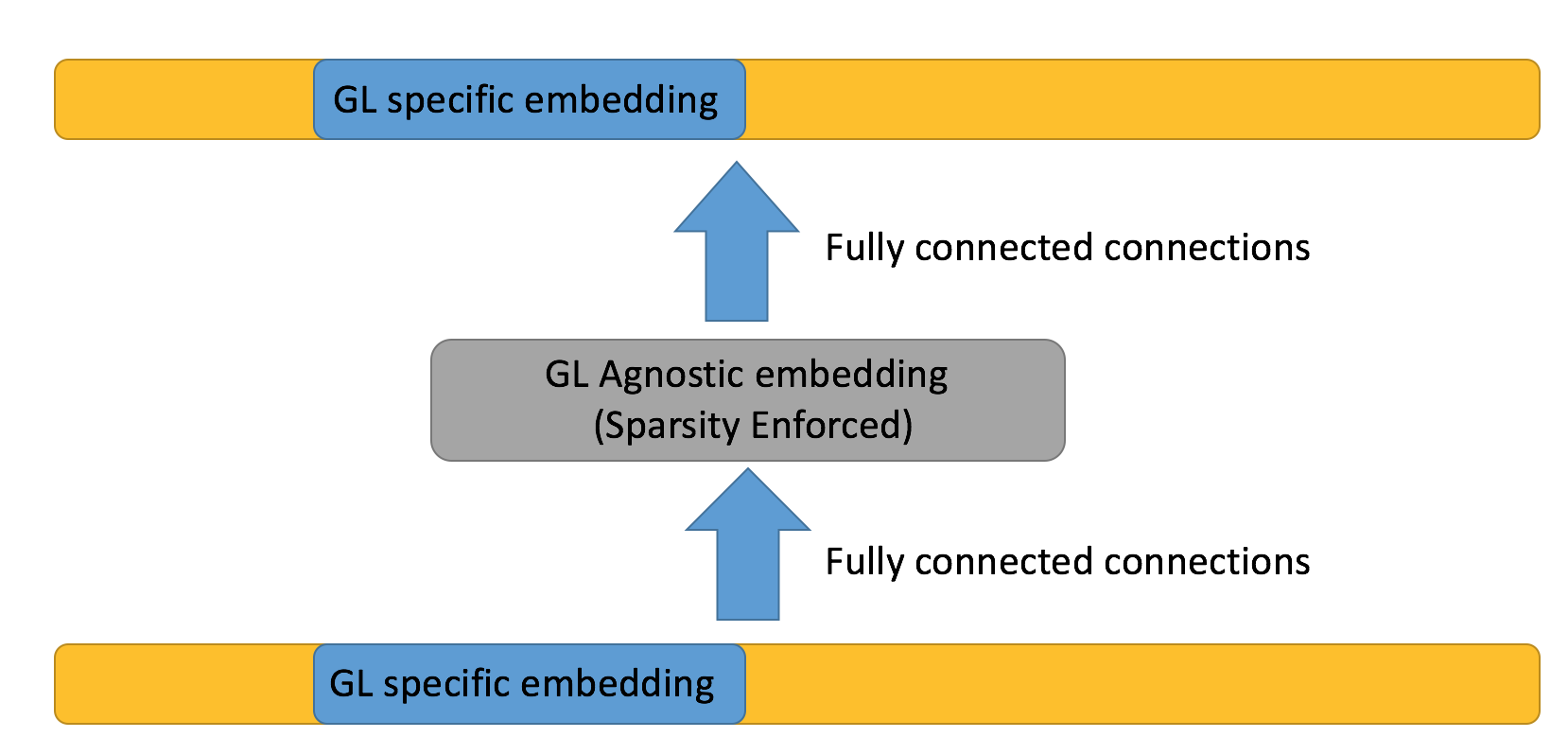}
        \caption{Sparse Autoencoder for PG-agnostic embeddings.}
        \label{autoencoder}
    \end{subfigure}
    \caption{Architecture of different components of MRNet-Product2Vec.}\label{fig:animals}
\end{figure}

The block diagram of MRNet-Product2Vec is shown in Fig. \ref{arch_1}. The word2vec representation of each word in a product title is fed through a Bidirectional RNN layer containing LSTM cells. The hidden layer representation from the forward and backward RNNs are concatenated to create the product embedding which is used to predict multiple task labels as described above. The network is trained jointly with all of these tasks.

Let us assume that the word2vec representation of words in a product title with T words are denoted as \{$\mathbf{x}_1$, $\mathbf{x}_2$,..., $\mathbf{x}_T$\}. We use a Bidirectional RNN, which has a forward RNN and a backward RNN. Let us assume that $\mathbf{h}_t^f$ and $\mathbf{h}_t^b$ denote the hidden states of the forward and backward RNN respectively at time $t$. The recursive equations for the forward and the backward RNN are given by:
\begin{equation} \label{eq1}
\mathbf{h}_t^f=\phi(W^f\mathbf{x}_t+U^f\mathbf{h}_{t-1}^f)
\end{equation}
\begin{equation} \label{eq2}
\mathbf{h}_t^b=\phi(W^b\mathbf{x}_t+U^b\mathbf{h}_{t-1}^b)
\end{equation}
\noindent Where $W^f$ and $W^b$ are the feedforward weight matrices for the forward and backward RNNs respectively. $U^f$ and $U^b$ are the recursive weight matrices for the forward and backward RNNs respectively. $\phi$ is usually a nonlinearity such as tanh or RELU. We use $\mathbf{h}_T=[\mathbf{h}_T^f, \mathbf{h}_T^b]$, as the final hidden representation of a product after all words in a product title have been fed through both the forward and backward RNNs. RNNs are trained using Backpropagation Through Time (BPTT)~\cite{bptt}. Although RNNs are designed to model sequential data, it has been found that simple RNNs are unable to model long sequences because of the vanishing gradient problem \cite{vanishing_gradient}. Long Short Term Memory units~\cite{lstm} are designed to tackle this issue where along with the standard recursive and feed-forward weight matrices there are input, forget, and output gates, which control the flow of information and can remember arbitrarily long sequences. In practice, it has been observed that RNNs with LSTM units are better than traditional RNNs (Equation \ref{eq1} and \ref{eq2}). Hence, we use LSTM units in the forward and backward RNNs. We skip the details of LSTM units and suggest interested readers to look at this article\footnote{\url{http://colah.github.io/posts/2015-08-Understanding-LSTMs/}} for an intuitive explanation on LSTMs.

Suppose, we want to train our network with N different tasks. Out of the N tasks, $N^c$ are classification, $N^r$ are regression and $N^d$ are decoding, i.e., $N^c+N^r+N^d=N$. $\mathbf{l}_m^c$ denotes the loss of the m-th classification task, $\mathbf{l}_p^r$ denotes the loss corresponding to the p-th regression task and $\mathbf{l}_q^d$ denotes the loss of the q-th decoding task. The losses corresponding to all the tasks are normalized such that one task with a higher loss cannot dominate the other tasks. While training we optimize the following loss which is the sum of the losses of all N tasks.

\begin{equation} \label{eq3}
L=\sum_{m=1}^{m=N^c}{\mathbf{l}_m^c}+\sum_{p=1}^{p=N^r}{\mathbf{l}_p^r}+\sum_{q=1}^{q=N^d}{\mathbf{l}_q^d}
\end{equation}

The hidden representation $\mathbf{h}_T$ corresponding to a product is projected to multiple output vectors ($\mathbf{o}_n$ for $n$-th task) using task specific weights and biases (Equation \ref{eq4}). The loss is computed as a function of the output vector and the target vector according to the type of a task. For example, if the task is a five-way classification, $\mathbf{h}_T$ is projected to a five dimensional output, followed by a softmax layer and eventually a cross-entropy loss is computed between the softmax output and the true target labels. For the regression task, $\mathbf{h}_T$ is projected to a scalar and a squared loss is computed with respect to the true score. Similarly, in the decoding task, $\mathbf{h}_T$ is projected to a $5000$ dimensional vector $\mathbf{o}_n$ and a squared loss is computed between the projected representation and the target $5000$ dimensional TF-IDF representation. 

\begin{equation} \label{eq4}
\mathbf{o}_n= W_n\mathbf{h}_T+\mathbf{b}_n
\end{equation}

\noindent
\underline{\bf Optimization in Deep Multi-task Neural Network: } The cost function in Equation (\ref{eq3}) can be optimized in two different ways:
\begin{itemize}[noitemsep, leftmargin=*, topsep=0pt]
\item {\bf Joint Optimization:} At each iteration of training, update all weights of the network using gradients computed with respect to the total loss as defined in Equation (\ref{eq3}). However, if each training example does not have labels corresponding to all the tasks, training in this way may not be possible.
\item {\bf Alternating Optimization: } At each iteration of training, randomly choose one of the tasks and optimize the network with respect to the loss corresponding to that task only. In this case, only the weights which correspond to that particular task and the weights of task-invariant layers (the Bidirectional RNN in our case) are updated. This style of training is useful when we do not have all task labels for a product. However, the training might be biased towards a specific task if the number of training examples corresponding to that task is significantly higher than other tasks. 
\end{itemize}
While we were training MRNet, it was difficult to obtain all task labels for each product. Hence, alternating optimization was an obvious choice for us. We sample training batches from each task uniformly to avoid biasing towards any specific task.  

\subsection{Product Group (PG) Agnostic Embeddings}
\label{gl_agnostic}

While training the proposed network, we trained a separate model for each PG because the label distribution could be very different across PGs. For example, the median price and the price range of jewelry is very different from that of books. Similarly, the materials used in clothes (cotton, polyester etc.) are different from that of kitchen items which are usually made of aluminium, metal or glass. If we train one model across all PGs, the embeddings are unlikely to capture the finer intra-PG variations in their representations. Hence, we build one model for each PG. In this paper, we train 23 different models for 23 different PGs.

The PG-specific embeddings can be used for any ML problem which is either PG-specific (all train and test data are from a particular PG) or when there are a large number of training examples from each PG such that separate PG-specific models can be trained. However, in many practical situations, none of the above might be true. Hence, it is also important to have product embeddings which are PG-agnostic such that ML models can be trained with products spanning multiple PGs. We handle this problem by training a sparse autoencoder~\cite{sparse_autoencoder_1} that projects the PG-specific embeddings to a PG-agnostic space (Fig. \ref{autoencoder}).

Let us assume that each PG-specific embedding has a dimension $d$ and there are total $G$ ($23$ in our case) PG-specific embeddings. First, we represent the embeddings from PG $g$, using a $Gd$ dimensional vector, where the PG-specific embedding is placed at the index range $(g-1)d+1:gd$ and the rest are filled with zeros. This vector is called $\mathbf{x}_{ga}$. We train a sparse autoencoder where we reconstruct $\mathbf{x}_{ga}$ through a fully connected network containing a hidden layer of dimension $2d$. The hidden layer representation is used as the PG-agnostic embedding. We enforce sparsity such that the autoencoder can learn interesting structures from the data and does not end up learning an identity function. 

\section{Experimental Results}
\label{results}

In this section, we evaluate MRNet-Product2Vec in various ways. In Sect. \ref{quantitative_analysis},  we discuss the quantitative results while qualitative studies are discussed in Sect. \ref{qualitative_analysis} and Sect. \ref{feat_interpret}.

\noindent
\underline{\bf Architecture and Framework Details:} In MRNet-Product2Vec, there is one layer of Bidirectional RNN containing LSTM nodes followed by multiple classification/regression/decoding units. We train each PG-specific model with maximum one million training samples from a PG corresponding to each training task. It took around 30 minutes to train each epoch using one Grid K520 GPU. While training the PG-agnostic embeddings, we used 500K randomly chosen products from each PG (total 11.5 million for 23 PGs). The sparse autoencoder took around 20 minutes per epoch while training. 

\subsection{Quantitative Analysis}
\label{quantitative_analysis}

Product embeddings can be created in many possible ways. They can capture different kinds of signals about products and can have varying performance in different end-tasks. To get a sense of the efficacy of MRNet-Product2Vec, we consider five different classification tasks. These tasks are different from the tasks that are used to train MRNet-Product2Vec. 
\begin{enumerate}[noitemsep, leftmargin=*]
\item {\bf Plugs:} In this binary classification problem, the goal is to predict if a product has an electrical plug or not.  This dataset has 205,535 labeled products. Here we perform five-fold cross validation and report the average AUC.
\item {\bf SIOC:} This classification problem tries to predict if a product can ship in its own container (SIOC) provided by the seller or if the e-commerce company needs to provide an additional container for this product. This is also a binary classification problem. This dataset has 296,961 labeled examples. Five-fold cross validation is performed and the average AUC over five-folds is reported.
\item {\bf Browse Categories:} This is a multi-class classification problem where products from the PG toy are classified into 75 different website browse categories (e.g.: baby toys, puzzles, and outdoor toys). There are a total of 150,197 samples in this dataset. We perform five-fold cross validation and report the average accuracy. 
\item {\bf Ingestible Classification:} We apply MRNet-Product2Vec on a product classification problem which predicts if a product is ingestible or not. However, only $1500$ training samples are available for learning a classifier. We perform five-fold cross validation and report the average AUC over five-folds.
\item {\bf SIOC (unseen population):} We believe that if the test data distribution is significantly different from the train data distribution, dense embeddings such as MRNet-Product2Vec should perform better than sparse/high-dimensional TF-IDF representations. We simulate that by modifying the SIOC dataset using the following steps. First, we split the full dataset into fixed training and test parts. We filter out each test product for which the maximum intersection of it's title with any training product title is larger than a threshold $t_h$. All the remaining products are used as the test data set. The lower the threshold, the difference between the test and the train data distribution increases. We fixed a training dataset of 150K examples and used $t_h=0.2$ with 271 test examples (106 +ve and 165 -ve). We report the AUC.
\end{enumerate}
We compare MRNet-Product2Vec with the sparse and high-dimensional TF-IDF representation of product titles for all the five classification tasks. The sparse TF-IDF representation for each classification task was created using only the training examples corresponding to that task. We use the PG-agnostic version of MRNet-Product2Vec (dimensionality: 256) for SIOC, PLUGs, ingestible and SIOC (unseen population) as the data spanned multiple PGs. We use the PG-specific version (dimensionality: 128) of MRNet-Product2Vec for browse category classification as all the samples were from the same PG, i.e., toy. We use MRNet-Product2Vec and TF-IDF in two different classifiers, Logistic Regression and Random Forest, and report the evaluation metric for both of them in Table \ref{benchmark}. We observe that on Plugs and SIOC, MRNet-Product2Vec is almost as good as sparse and high-dimensional TF-IDF in spite of having a much lower dimension than that of TF-IDF. However, on Browse Categories, MRNet-Product2Vec performs much worse than TF-IDF. This happens because out of the 15 tasks, only the subcategory classification task is somewhat related to browse categories. Hence, the browse category related information that is encoded in MRNet-Product2Vec is not sufficient enough for this ``hard'' 75-class classification task. MRNet-Product2Vec performs better than sparse and high-dimensional TF-IDF on ingestible and SIOC (unseen population). Since the dense embeddings are semantically more meaningful, i.e., it knows that a chair and a sofa are similar objects, they should be able to learn classifiers even from smaller training datasets (such as ingestible) and generalize well for unseen test population (SIOC with unseen population). However, sparse and high-dimensional TF-IDF is not as effective in these scenarios. Overall, we observe that MRNet-Product2Vec is mostly comparable to TF-IDF in spite of having less than 3\% of the TF-IDF dimension.

\begin{center}
\centering
\captionof{table}{Results on five classification tasks. RF: Random Forest, LR: Logistic Regression. TF-IDF dim.: >10K, MRNet-Product2Vec dim.: 256 and 128. All numbers are relative w.r.t TF-IDF-LR.}
\scriptsize
\label{benchmark}
\begin{tabular}{ |c|c|c||c|c| }
\hline
{\bf Task} & {\bf MRNet-RF} & {\bf MRNet-LR} & {\bf TF-IDF-RF} \\ \hline
\multirow{1}{*}{{\bf Plugs}} & -2.8\% & -9.72\% & -2.8\%  \\\hline
\multirow{1}{*}{{\bf SIOC}} & -5.81\% & -18.60\% & -9.3\%   \\\hline
\multirow{1}{*}{{\bf Browse Categories}} & -16.67\% & -26.38\% & -25.0\%   \\\hline
\multirow{1}{*}{{\bf Ingestible}} & 0\% & {\bf +2.15\%} & -11.8\%  \\ \hline
\multirow{1}{*}{{\bf SIOC (unseen)}} & {\bf +10\%} & 0\% & -3.33\%  \\ \hline
\end{tabular}
\end{center}

\subsection{Nearest Neighbor Analysis}
\label{qualitative_analysis}

\begin{figure}
    \centering
    \includegraphics[width=0.95\textwidth]{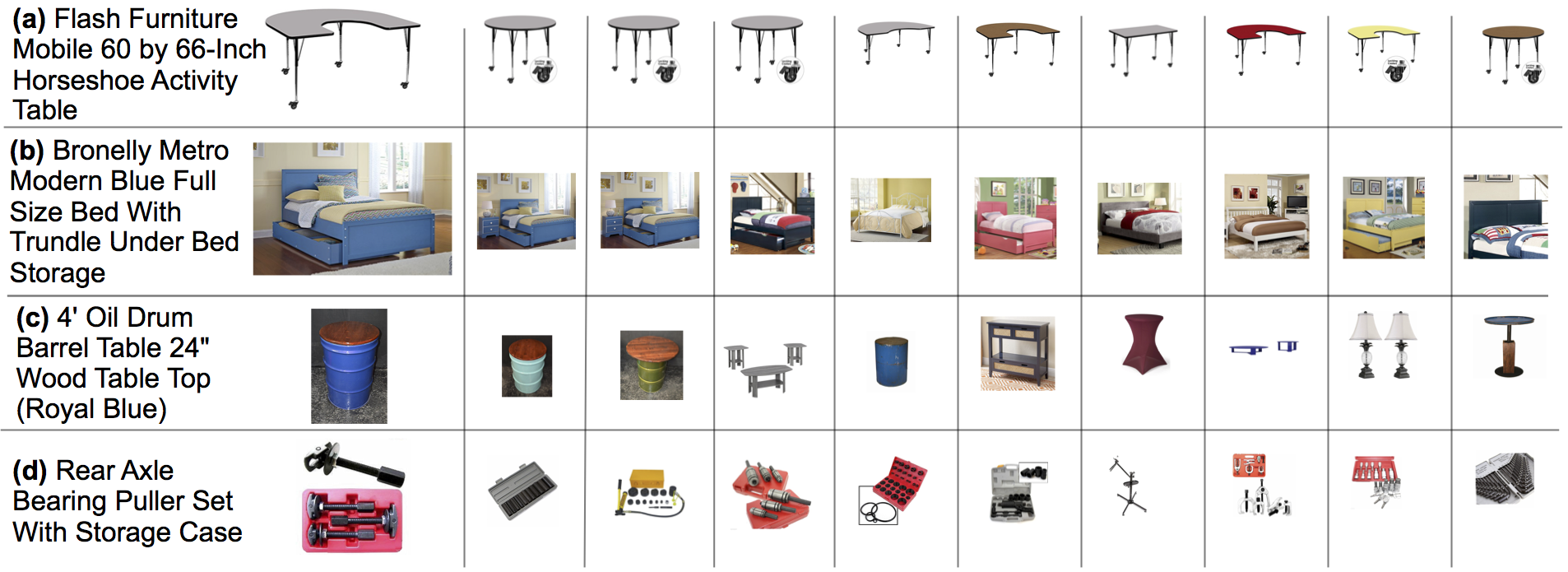}
    \caption{Nearest neighbors computed using MRNet-Product2Vec for each query product (first column) (best viewed in electronic copy).}
    \label{nearest_neighbor}
\end{figure}

We study the characteristics of MRNet-Product2Vec by analyzing the nearest neighbors (NN) of several products. Since computing meaningful NNs is not straightforward using the sparse TF-IDF features, we do not show the NNs using this method. We created a universe of 220K products from the PG furniture and computed the NNs of several randomly chosen products based on the Euclidean distance. In Fig. \ref{nearest_neighbor}, we show the first nine NNs of four hand-picked products. In (a), MRNet-Product2Vec finds several {\bf grey} colored tables as NNs. In (b), several {\bf full-sized} beds are obtained as NNs. In (c), MRNet-Product2Vec fetches four {\bf blue-colored} tables and two ``drum barrel" tables as NNs. In (d), MRNet-Product2Vec produces several {\bf tools/tool-boxes} as NNs. Overall, we can see that MRNet-Product2Vec has learned several intrinsic characteristics of products such as size, color, type etc, which were used to train MRNet-Product2Vec.

\subsection{MRNet-Product2Vec Feature Interpretation}
\label{feat_interpret}

MRNet-Product2Vec is trained with multiple tasks to incorporate different product related signals. We performed some preliminary analysis on the PG-agnostic embeddings (256 dimensional) to detect if a subset of features encode a particular signal (such as size, weight, electrical properties etc.). First, MRNet-Product2Vec is used for the battery classification task (one of our training tasks) and multiple Random Forest (RF) models with randomly chosen subsets of the training data are built. We found out that there are $29$ features that always appear in the top quartile (64) of all the features with respect to RF feature importance. That indicates that these $29$ features in MRNet-Product2Vec are indicative of product's electrical properties and some of these should play a role in plugs classification (evaluation task). Indeed, we find that there are $28$ features which are important in the context of plugs classification and $8$ of these features were also important in the battery classification task. Likewise, we find that $13$ important features for the weight classification training task are also important for the SIOC evaluation task. This demonstrates that MRNet-Product2Vec encodes different product characteristics which play a crucial role in the final evaluation tasks.

\section{Language Agnostic MRNet-Product2Vec}
\label{language_agnostic}

E-commerce companies usually sell products across multiple countries and language-regions. There are many scenarios when it is important to compare products that have details such as title, description, and bullet-points in different languages. When a seller lists a product in a country, the e-commerce company would like to know if that product is already listed in other countries for accurate stock-accounting and price estimation. A customer from UK might like to know if a product which she liked in the France website is available for purchase from the UK website or not. Often it is also required to apply a trained ML model from a particular language-region to a different language-region product because labeled data in that language may not be available. For each of these use-cases, it is important to learn cross-language transformations, such that products from different countries/language-regions can be compared seamlessly. To address this issue, we propose to use a variant of multimodal autoencoder~\cite{ngiam2011multimodal} that can project MRNet-Product2Vec trained in different languages to a common space for comparison. 

\begin{figure}
    \centering
    \begin{subfigure}[b]{0.51\textwidth}
        \includegraphics[width=\textwidth]{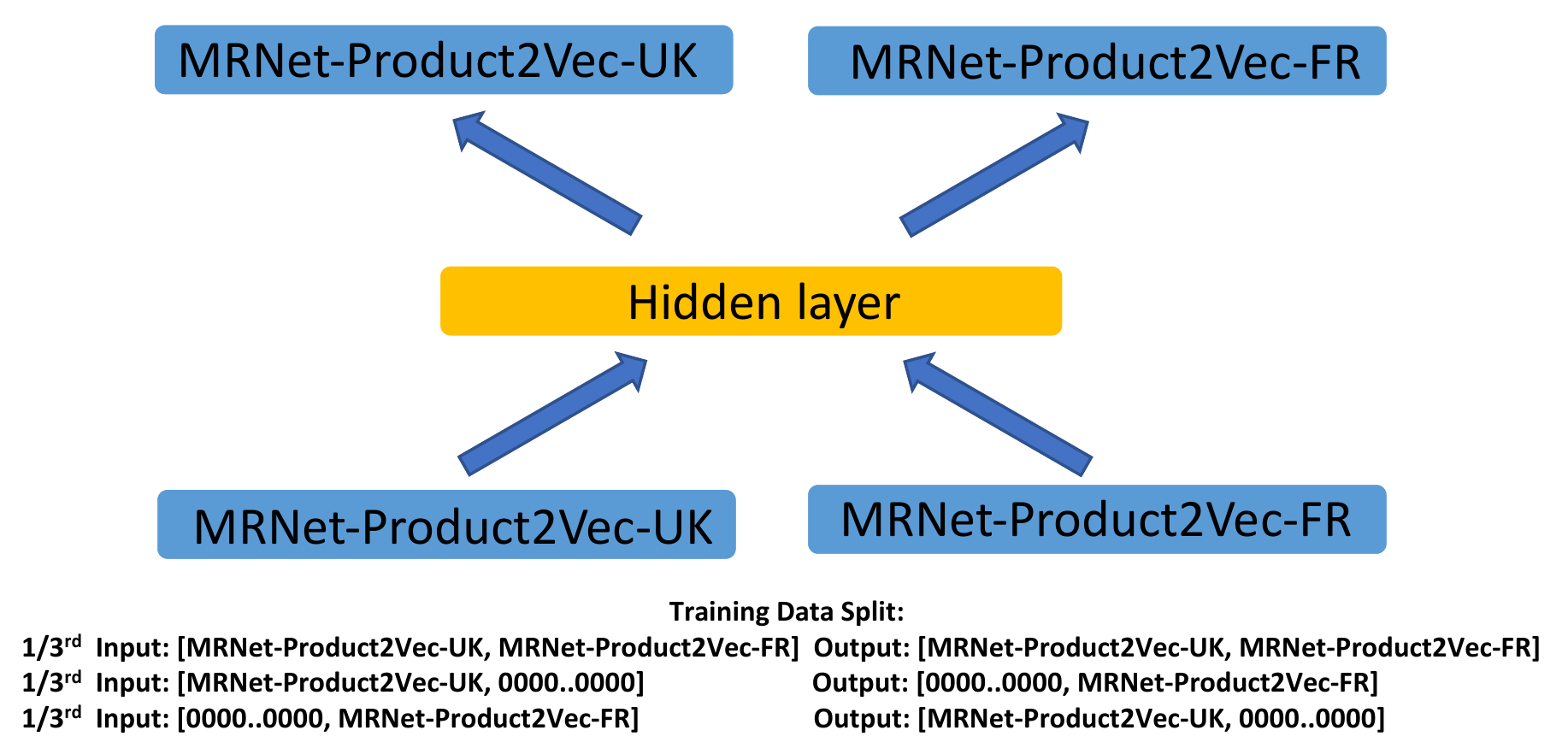}
        \caption{Architecture of Multimodal Autoencoder}
        \label{arch_mm}
    \end{subfigure}
    ~ 
    \begin{subfigure}[b]{0.46\textwidth}
        \includegraphics[width=\textwidth]{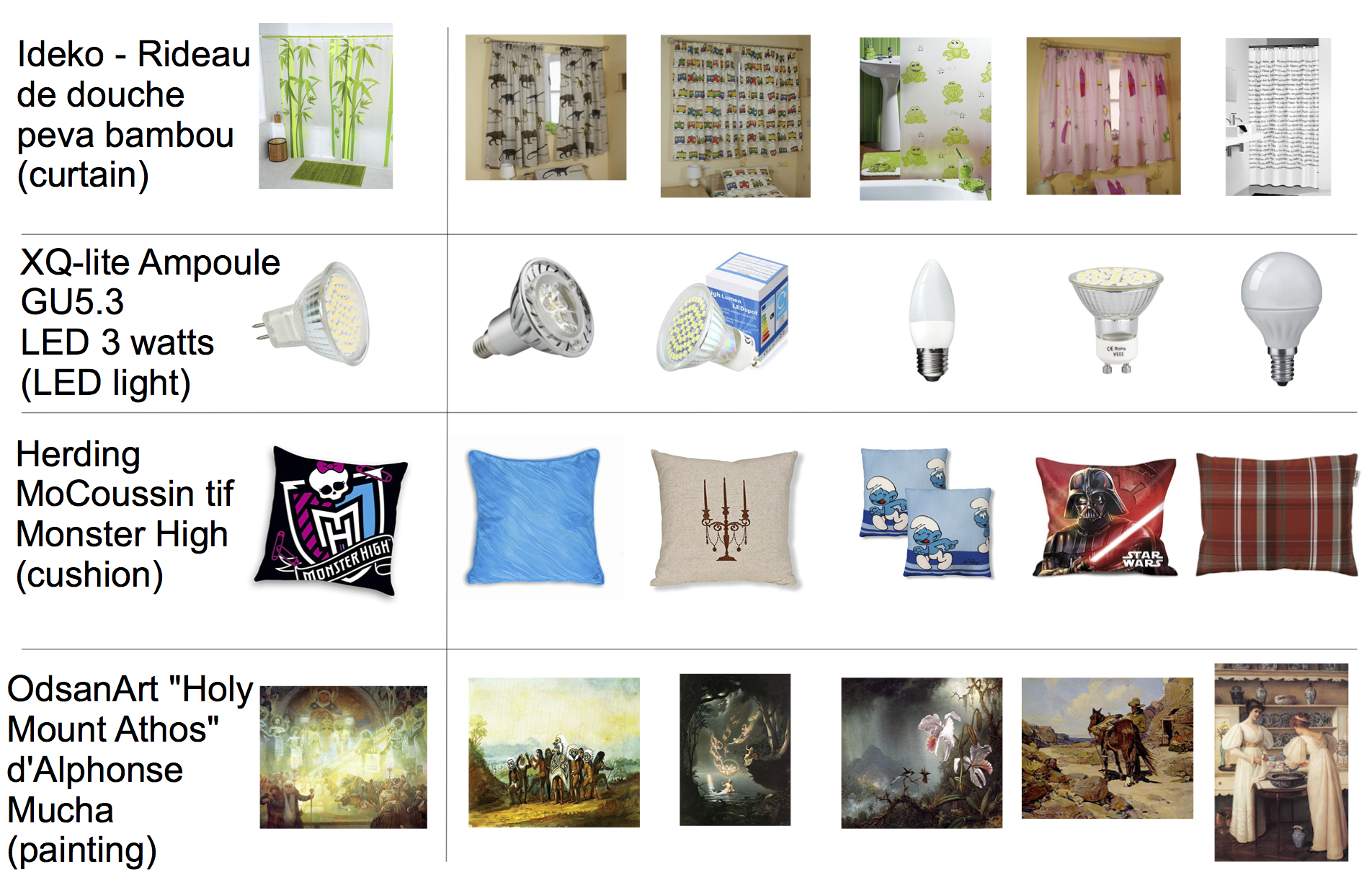}
        \caption{Nearest neighbors from UK products w.r.to a French product.}
        \label{la_mrnet_nn}
    \end{subfigure}
    \caption{Language Agnostic MRNet-Product2Vec (best viewed in electronic copy).}\label{fig:LA-mrnet}
\end{figure}

Let us assume that we have $P$ paired product titles from two different countries UK and France, i.e., for each product title in French, we know the corresponding UK title and vice versa. We separately train MRNet-Product2Vec for UK-english and French and refer them as MRNet-Product2Vec-UK and MRNet-Product2Vec-FR respectively. MRNet-Product2Vec-UK and MRNet-Product2Vec-FR are used to obtain the embeddings for $P$ products. The corresponding embeddings for the p-th product are defined as $\mathbf{x}_p^{UK}$ (dim. 256) and $\mathbf{x}_p^{FR}$ (dim. 256) respectively. Now, we train an autoencoder (Fig. \ref{arch_mm}) which has input $\mathbf{x}_p$ (dim. 512), output $\mathbf{y}_p$ (dim. 512) and a hidden layer of dimension 256. Let us assume that $\mathbf{0}$ denotes a zero vector of 256 dimension. The training data for this network consists of three parts: (1) $\mathbf{x}_p=[\mathbf{x}_p^{UK}, \mathbf{0}]$ and $\mathbf{y}_p=[\mathbf{0},\mathbf{x}_p^{FR}]$, (2) $\mathbf{x}_p=[\mathbf{0},\mathbf{x}_p^{FR}]$ and $\mathbf{y}_p=[\mathbf{x}_p^{UK}, \mathbf{0}]$ and (3) $\mathbf{x}_p=[\mathbf{x}_p^{UK}, \mathbf{x}_p^{FR}]$ and $\mathbf{y}_p=[\mathbf{x}_p^{UK},\mathbf{x}_p^{FR}]$. We train the autoencoder with batches of size 256, where each batch is randomly selected from the full training data. The trained network is used to project a product's MRNet-Product2Vec-FR to the corresponding MRNet-Product2Vec-UK space. The projected MRNet-Product2Vec-UK representation is used to find the nearest UK products corresponding to a French product. We demonstrate a few French products and their UK nearest neighbors in Fig. \ref{la_mrnet_nn}. We could have also projected the UK products to the French embedding space or project both of them to the common shared space for comparison. Although the results are preliminary, this demonstrates that we can use a multimodal autoencoder to effectively compare embeddings from different language-regions.

\vspace{-2mm}

\section{Discussion and Future Work}

In this paper, we propose a novel variant of e-commerce product embeddings called MRNet-Product2Vec, where different product related signals are explicitly injected into their embeddings by training a Discriminative Multi-task Bidirectional RNN. Initially, PG-specific embeddings are learned and then a PG-agnostic version is learned using a sparse autoencoder. We evaluate the proposed embeddings qualitatively and quantitatively and establish it's effectiveness. We also propose a multimodal autoencoder for comparing products across different countries (i.e., languages) and provide initial results using that. MRNet-Product2Vec has been applied to generate product embeddings of around 2 billion products and have been made available internally within our company for product related ML model building. We periodically retrain MRNet to keep the model updated with the dynamic signals and also update the resulting embeddings of all products. We note that MRNet-ProductVec is suitable for cold-start scenarios as the embeddings can be created using only product titles, which are available as part of the catalog data. Although MRNet-Product2Vec has been trained with the proposed set of 15 different tasks, the framework provides the flexibility to learn embeddings with any other set of tasks or fine-tune the already learnt embeddings with additional tasks.

\bibliography{mrnet}{}
\bibliographystyle{splncs03}

\newpage
\appendix

\end{document}